\documentclass[runningheads]{llncs}

\usepackage{eccv}

\usepackage{eccvabbrv}
\usepackage{float}

\usepackage{graphicx}
\usepackage{booktabs}
\usepackage{bm}

\usepackage[accsupp]{axessibility}  

\usepackage{hyperref}

\usepackage{orcidlink}
\usepackage{amsmath}
\usepackage{multirow}
\usepackage{colortbl}
\usepackage{algorithm}
\usepackage{algpseudocode}
\usepackage{amssymb}
\usepackage[table]{xcolor}

\begin{document}

\title{AdvSplat: Adversarial Attacks on Feed-Forward Gaussian Splatting Models} 

\titlerunning{Abbreviated paper title}

\author{Yiran Qiao \and
Yiren Lu \and
Yunlai Zhou \and Rui Yang \and Linlin Hou \and Yu Yin \and Jing Ma}

\authorrunning{F.~Author et al.}

\institute{Case Western Reserve University, Cleveland OH 44106, USA \\
\email{\{yxq350, yxl3538, yxz3057, rxy337, lxh663, yxy1421, jxm1384\}@case.edu}\\}

\maketitle

\begin{figure}
  \includegraphics[width=\textwidth]{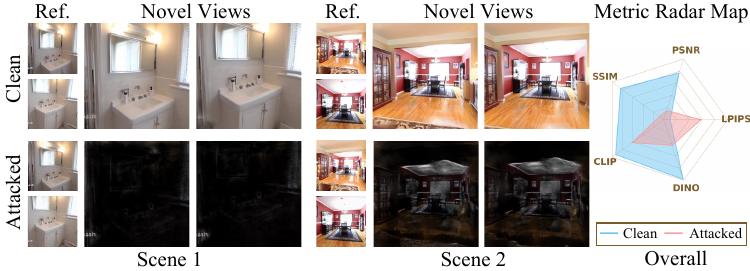}\\
  \centering
  \caption{White-box attack (here we use PGD attack \cite{madry2017towards}) results against feed-forward 3DGS model (here we use NoPoSplat\cite{ye2024no})) on the RE10K dataset. "Ref." denotes two input reference views, and "Novel Views" denote the newly rendered target viewpoints. The first row shows results on clean inputs, while the second row shows results on adversarially perturbed inputs. The rightmost radar chart shows the rendered image quality with clean/attacked inputs, where PSNR is normalized with 30 as the maximum value. As shown, both quantitative and qualitative results consistently indicate that even imperceptible perturbations can severely degrade the reconstruction quality.}
  \label{fig:teaser}
\end{figure}

\begin{abstract}
3D Gaussian Splatting (3DGS) is increasingly recognized as a powerful paradigm for real-time, high-fidelity 3D reconstruction. However, its per-scene optimization pipeline limits scalability and generalization, and prevents efficient inference. Recently emerged feed-forward 3DGS models address these limitations by enabling fast reconstruction from a few input views after large-scale pretraining, without scene-specific optimization. Despite their advantages and strong potential for commercial deployment, the use of neural networks as the backbone also amplifies the risk of adversarial manipulation. In this paper, we introduce AdvSplat, the first systematic study of adversarial attacks on feed-forward 3DGS. We first employ white-box attacks to reveal fundamental vulnerabilities of this model family. 
We then develop two improved, practically relevant, query-efficient black-box algorithms that optimize pixel-space perturbations via a frequency domain parameterization: one based on gradient estimation and the other gradient-free, without requiring any access to model internals. Extensive experiments across multiple datasets demonstrate that AdvSplat can significantly disrupt reconstruction results by injecting imperceptible perturbations into the input images. Our findings surface an overlooked yet urgent problem in this domain, and we hope to draw the community’s attention to this emerging security and robustness challenge.
\end{abstract}

\section{Introduction}
\label{sec:intro}

3D reconstruction recovers 3D geometry and structure from multi-view 2D images. Representative paradigms include NeRF\cite{mildenhall2021nerf} and 3D Gaussian Splatting (3DGS)\cite{kerbl20233d}, the latter becoming dominant due to real-time rendering and high-fidelity reconstruction. These strengths have driven 3DGS adoption in robotics\cite{keetha2024splatam,lu2024manigaussian,matsuki2024gaussian,zou20253d}, autonomous driving\cite{hess2025splatad,khan2025autosplat,zhou2024drivinggaussian}, and content generation\cite{xiang2025native,xiang2025structured,hunyuan3d2025hunyuan3d,bahmani2025lyra,wu2025direct3d}. However, conventional 3DGS relies on per-scene optimization, limiting inference efficiency, scalability, and generalization, and often underperforming in sparse-view settings\cite{wu2025difix3d+}. These limitations hinder its practicality in few-view, near-instant reconstruction scenarios\cite{liu2025monosplat,charatan2024pixelsplat}. In contrast, \textit{feed-forward 3DGS}\cite{szymanowicz2024splatter,charatan2024pixelsplat} alleviates these issues through large-scale pretraining with few-view inputs\cite{zhang2025advances}, offering strong potential for foundation-model-style 3D reconstruction and broader commercial deployment.

\looseness -1
Given these desirable properties and strong potential of feed-forward 3DGS, ensuring its robustness becomes critically important. While a few studies\cite{lu2024poison,zeybey2024gaussian} have revealed the vulnerability of conventional 3DGS by designing adversarial attacks, the vulnerability of feed-forward 3DGS remains largely unexplored. This raises an overlooked yet crucial question, one that we aim to answer in this paper: \textit{How robust are feed-forward 3DGS models against adversarial attacks, and how can they be effectively attacked?} Although white-box attacks are simple and effective, their applicability in real-world scenarios is limited by the requirement of access to model internals. Motivated by this, we focus on the more practical \textit{black-box attack setting}.  This problem remains uninvestigated in the community, especially due to two key challenges. 
\textbf{Challenge 1: unknown model parameter weights.} In practice, the internal weights of commercial models are typically unavailable. Instead, access is restricted to API queries, \ie, only the model’s inputs and outputs can be observed. This makes the design of effective attacks substantially more challenging, especially for feed-forward 3DGS models which are often large-scale.
\textbf{Challenge 2: high computational costs. } Traditional black-box attacks are typically categorized into transfer-based and query-based attacks. Feed-forward 3DGS models often exhibit substantially larger feature-space discrepancies than conventional classifiers, largely because they are built upon different high-capacity transformer backbones\cite{wang2025vggt,yang2024depth} with distinct pretraining data, architectural biases, and representation geometries, making transfer-based methods almost impossible. Query-based attacks can be further divided into non-optimization\cite{andriushchenko2020square,guo2019simple,brendel2017decision} and optimization-based\cite{ilyas2018black} approaches. Non-optimization attacks often rely on decision-boundary exploration or random noise search, but both are ineffective for feed-forward 3DGS models: these models lack a decision boundary since their tasks are not classification, and their large capacity makes them relatively insensitive to random noise. We therefore focus on optimization-based black-box attacks, which optimize adversarial perturbations through model queries. However, despite faster inference compared to traditional 3DGS, feed-forward 3DGS models remain large, making the per-query cost a bottleneck for query-intensive attacks. Moreover, reconstruction outputs lie in a high-dimensional pixel space, requiring extensive queries to explore adversarial perturbations, further exacerbating this challenge.

In this paper, we \textbf{answer the research question of feed-forward 3DGS robustness and design effective adversarial attacks against them}. 
As an initial experiment to reveal the model vulnerability, we applied representative white-box attacks (in particular, Projected Gradient Descent (PGD) \cite{madry2017towards}) designed for conventional classifiers to attack feed-forward 3DGS models. Unlike traditional 3DGS, the neural network architecture of feed-forward 3DGS makes it possible to leverage gradient information to craft adversarial perturbations on the input images. By constructing subtle perturbations through maximizing the reconstruction loss, we observe a significant degradation in rendering quality, as shown in \cref{fig:teaser}. This reveals a potential risk:\textit{ feed-forward 3DGS models are not inherently robust, and an attacker can severely corrupt the model’s outputs by introducing imperceptible perturbations}. 
Inspired by this, we design black-box attacks via optimizing subtle yet misleading perturbations through repeated queries to the feed-forward 3DGS model, either in a  \textit{gradient-based} or a \textit{gradient-free} manner (addressing Challenge 1). To reduce the number of queries and improve efficiency, we devise a Discrete Cosine Transform (DCT) based approach. Specifically, we first partition the image into equal-sized blocks and apply DCT to each block to transform it into the frequency domain. We then perturb only the low-frequency components, whose dimensionality is far lower than that of the pixel space, and transform them back to the image space via the inverse DCT (addressing Challenge 2). 

Given the vulnerabilities shown above, a highly plausible real-world scenario is that an adversary can repeatedly query a feed-forward 3DGS model to generate adversarially perturbed images and subsequently upload them to the internet. Users may download such data without noticing any abnormality. Once these attacked inputs lead to low-quality reconstructions, both the company’s commercial reputation and user trust may be adversely affected. Our work identifies and systematically investigates this issue, and draws the efficient and generalizable 3D reconstruction community’s attention to security and robustness. The contributions of this work can be concluded in threefold:
\begin{itemize}
    \item To the best of our knowledge, this work is the first to investigate a previously overlooked yet important security issue in the deployment of feed-forward 3DGS models, revealing their vulnerabilities to adversarial attacks.
    \item This paper proposes two black-box attack algorithms for feed-forward 3DGS, including a gradient-based and a gradient-free method, both operating in the frequency domain and requiring access only to model inputs and outputs.
    \item Extensive experiments across multiple state-of-the-art feed-forward 3DGS models and widely used datasets demonstrate that the proposed methods consistently achieve successful attacks.
\end{itemize}

\section{Related Work}

\subsection{3D Gaussian Splatting}
3DGS\cite{kerbl20233d} has recently emerged as a prevailing framework for 3D scene representation. It models geometry, appearance, and texture using an explicit set of anisotropic 3D Gaussian ellipsoids parameterized by their positions, scales, orientations, and colors. Unlike implicit approaches such as NeRF\cite{mildenhall2021nerf}, which heavily queries an MLP-parameterized radiance field, 3DGS enables real-time rendering with high visual fidelity. Moreover, its explicit representation offers improved interpretability and scalability, and is more amenable to vision and graphics applications, including virtual reality\cite{jiang2024vr,tu2024fast}, dynamic scene reconstruction\cite{yang2023real,yang2024deformable,wu20244d}, open-vocabulary 3D semantic segmentation\cite{wu2024opengaussian,qin2024langsplat,shi2024language,ye2024gaussian,lu2025segment}, and 3D editing\cite{qu2025drag,dong20263dgs,yan20243dsceneeditor}. However, the per-scene optimization characteristic of 3DGS limits its generalization and scalability, and also makes it ill-suited for real-time inference requirements.

\subsection{Feed-Forward 3D Gaussian Splatting}
Feed-forward 3DGS models alleviate the above limitations by pretraining on large-scale datasets\cite{zhang2025advances}. Given only a small set of input images, they predict the parameters of 3D Gaussians with a single forward pass of a network, enabling fast inference. Benefiting from large-scale training data, they also exhibit strong novel-view synthesis capability. For simplicity and clarity, we categorize feed-forward 3DGS methods into pose-known and pose-free approaches.
\subsubsection{Pose-Known Methods}
PixelSplat\cite{charatan2024pixelsplat} leverages an epipolar transformer to infer a scene-specific scale factor and estimates a probabilistic depth distribution, which is used to determine 3D Gaussian positions. Building on this, LatentSplat\cite{wewer2024latentsplat} adopts a generative approach to encode uncertainty in the latent space, improving reconstruction quality in high-uncertainty regions. A key limitation of these methods is the unreliability of mapping image features to depth. To address this issue, MVSplat\cite{chen2024mvsplat} builds a cost-volume representation via plane sweeping and leverages cross-view feature similarities to improve depth estimation. In comparison, DepthSplat\cite{xu2025depthsplat} leverages pretrained monocular depth features to improve 3D reconstruction.

\subsubsection{Pose-Free Methods}
In practice, obtaining accurate camera poses is often non-trivial. Consequently, a growing line of recent work explores pose-free approaches. NoPoSplat\cite{ye2024no} adopts the input-view camera coordinate system as a canonical space and predicts Gaussian primitives without requiring ground-truth poses. SelfSplat\cite{kang2025selfsplat} performs pose-free, generalizable 3DGS by coupling explicit Gaussians with self-supervised depth and pose estimation. In contrast, AnySplat\cite{jiang2025anysplat} predicts Gaussians and poses simultaneously and uses differentiable voxelization to aggregate pixel-wise primitives into voxel-wise Gaussians for efficiency.

\subsection{Adversarial Attacks}

Adversarial attacks are methods for generating adversarial examples. For a conventional classification network, an adversarial example is obtained by adding imperceptible, non-random perturbations to the original input, yielding a new input that can arbitrarily change the model’s prediction\cite{szegedy2013intriguing}. Such attacks are typically studied under two settings: white-box and black-box. White-box attacks craft perturbations by leveraging access to the model’s gradients\cite{moosavi2016deepfool,goodfellow2014explaining,madry2017towards}. In contrast, black-box attacks primarily rely on transferability\cite{liu2016delving} or query-based optimization\cite{ilyas2018black,andriushchenko2020square,guo2019simple,brendel2017decision}. With the rise of 3DGS, its security has gradually attracted increasing attention. PoisonSplat\cite{lu2024poison} formulates a bilevel optimization problem to design a computation-cost attack, thereby exposing the vulnerability of 3DGS systems. However, security and robustness analyses for feed-forward 3DGS models are still absent, and our work fills this gap.

\section{Method}

\begin{figure}[t]
  \centering
  \includegraphics[width=\textwidth]{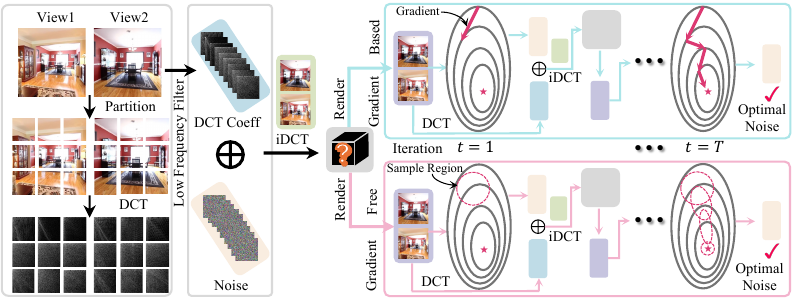}
  \caption{Pipeline of our proposed method, including gradient-based and gradient-free variants. The \colorbox[HTML]{BFDCE7}{blue blocks} denote the DCT coefficient maps, the \colorbox[HTML]{F9EDE0}{yellow blocks} denote the sampled noise, the \colorbox[HTML]{D9E4C2}{green blocks} denote the images after iDCT, and the \colorbox[HTML]{C6C4DF}{purple blocks} denote the rendered images. The black box in the center is the feed-forward 3DGS model with unknown parameters, illustrated by the \colorbox[HTML]{D9D9D9}{gray block}. The gray ellipses depict a contour map of the loss function, and the red star indicates the maximum. The red arrow inside the ellipse shows the gradient direction, while the red dashed ellipse indicates the noise sampling region. Finally, the noise sample marked with a red checkmark is the optimal perturbation we seek.}
  \label{fig:pipeline}
\end{figure}

\looseness -1
We first reveal the vulnerability of feed-forward 3DGS models via white-box attacks using PGD (assuming access to the weights). We then focus on black-box attacks and propose two improved, effective, and query-efficient attack algorithms (gradient-based and gradient-free) based on an efficient frequency domain parameterization of perturbations, as shown in Fig. \ref{fig:pipeline}. 

\subsection{ Preliminaries}
\subsubsection{Feed-Forward 3DGS}
Suppose we are given $N$ input images $\{ \bm{I}^i \}_{i=1}^N$ ($\bm{I}^i \in \mathbb{R}^{H \times W \times 3}$, $H$ is image height and $W$ is image width) along with their corresponding camera projection matrices $\{ \bm{P}^i \}_{i=1}^N$ (calculated via intrinsic $\bm{K}^i$, rotation $\bm{R}^i$ and translation $\bm{t}^i$: $\bm{P}^i=\bm{K}^i[\bm{R}^i|\bm{t}^i]$, and can be empty for pose-free settings). The goal of feed-forward 3DGS is to learn a mapping $\bm{f_\theta}$ from input images to the parameters of 3D Gaussians:
\begin{equation}
    \bm{f_{\theta}}:\{(\bm{I}^i,\bm{P}^i)\}_{i=1}^{N}\mapsto \{(\bm{\mu}_j,\alpha_j,\bm{\Sigma}_j,\bm{c}_j)\}_{j=1}^{H\times W\times N},
\end{equation}
where $\bm{\theta}$ denotes the parameters of the feed-forward network; $\bm{\mu}_j$ is the Gaussian mean; $\alpha_j$ is the opacity; $\bm{\Sigma}_j$ is the covariance; and $\bm{c}_j$ denotes the spherical harmonics coefficients. Given a specific camera pose, each pixel $(p,q)$ color $\bm{I}
_r(p,q)$ is calculated via alpha blending:
\begin{equation}
    \bm{I}_r(p,q) = \sum_{j \in \mathcal{V}} c_j \alpha_j \prod_{l=1}^{j-1} (1 - \alpha_l),
\end{equation}
where $\mathcal{V}$ is a set of visible Gaussians that contribute to the pixel.
\subsubsection{PGD Attack}
PGD is a white-box adversarial attack algorithm designed for a traditional classifier that iteratively updates an input image using the gradient of the loss $\mathcal{L}$ with respect to the input $x$, seeking a perturbation that most degrades the model’s decision. Starting from the clean image, PGD performs multiple update steps; after each step $t$, the intermediate adversarial sample $x_t$ is projected back into an $\ell_\infty$ constraint set centered at the original input with a specified radius. The procedure is summarized as follows:
\begin{equation}
\label{eq1}
x_{t+1} = \Pi_{x + \mathcal{S}} \left( x_t + \eta \, \text{sgn} \left( \nabla_x \mathcal{L}(\theta, f(x_t), y) \right) \right),
\end{equation}
where $y$ is the label, $\Pi$ is the projection operator, $\mathcal{S}$ defines the radius of the $\ell_\infty$-norm ball, $\eta$ denotes the step size, sgn$(\cdot)$ is the sign function, and $\theta$ represents the parameters of the target model.

\subsection{Vulnerability of Feed-Forward 3DGS Model}
Similar to classification models, exposing the vulnerability of feed-forward 3DGS can be cast as an equivalent objective: whether one can construct imperceptible perturbations on the input images that substantially degrade reconstruction quality. To achieve this goal, we assume full access to the weights of the feed-forward 3DGS model. We then employ PGD, as shown in \cref{eq1}, to perform multi-step attacks and generate adversarial examples. Unlike classification, reconstruction does not involve discrete labels. Nevertheless, the attack objective is analogous: to increase the loss by crafting the manipulation of the input along the gradient direction. Instead of a classification loss, we use a standard combination of a photometric loss and an LPIPS loss as the overall objective. Although different methods employ different training losses, these two terms are almost universally adopted and typically contribute the most, making our choice both reasonable and broadly applicable. We formulate the white-box attacks on feed-forward 3DGS models as follows:
\begin{gather}
\label{pgd eq}
    \bm{I}_{t+1} = \Pi_{\bm{I} + \mathcal{S}} \left( \bm{I}_t + \eta \, \text{sgn} \left( \nabla_{\bm{I}} \mathcal{L}(\theta, f(\bm{I}_t), \bm{I}_t^r) \right) \right), \\
    \mathcal{L}=\text{MSE}(\bm{I}_t,\bm{I}_t^r)+\lambda\text{LPIPS}(\bm{I}_t,\bm{I}_t^r),
    \label{loss eq}
\end{gather}
where MSE stands for mean square error, $\bm{I}_t$ are input images at step $t$, $\bm{I}_t^r$ are rendered images at step $t$, with the same poses of $\bm{I}_t$, $\lambda$ is the balancing weight. The results in \cref{fig:teaser} show that white-box attacks are effective against feed-forward 3DGS models, producing adversarial examples that cause the reconstruction quality to collapse. This is also reflected consistently by the quantitative metrics in the radar map (more details in Appendix B), revealing the model’s inherent vulnerability.

\subsection{Black-Box Attack on Feed-Forward 3DGS Model}

\looseness -1
As discussed in \cref{sec:intro}, an adversarial example crafted on model A does not substantially affect feature extraction in model B. We validate this empirically, with results provided in Appendix C. Moreover, non-optimization query-based attacks are inherently ineffective against feed-forward 3DGS models. Based on this, we focus on optimization-based query attacks.

\subsubsection{Frequency-domain Parameterization of Perturbations}
\looseness -1
In query-based attacks, the required number of queries is positively correlated with the dimensionality of the attacked object: as the dimensionality increases, the number of explorable directions also grows, requiring more queries to distinguish favorable optimization directions from random search directions. Because input images in 3D reconstruction tasks are typically high-resolution, a large number of queries is required. This introduces several drawbacks. A larger number of queries substantially increases the computational overhead, thereby raising both the time and monetary costs of the attack. To reduce the number of queries and improve attack efficiency, we propose a frequency-domain parameterization of perturbations.

3D-GSW\cite{jang20243d} points out that the low-frequency components of an image contain the main structural information and can support robust information embedding. Inspired by this, we apply perturbations to the low-frequency components. We first divide the image into blocks $\bm{I_b}$ of equal size, each of size $n\times{n}$. Applying DCT in a block-wise manner provides a localized frequency representation, enabling finer spatial control and more efficient manipulation of region-specific image structures. In this work, we use the orthonormal DCT-II transform matrix $\bm{C}\in\mathbb{R}^{n\times n}$. The $(k,i)$-th entry $\bm{C}_{k,i}$ specifies the contribution of the input sample at spatial index $i\in\{0,1,\ldots,n-1\}$ to the DCT coefficient at frequency index $k\in\{0,1,\ldots,n-1\}$ (with $k=0$ corresponding to the DC component), and is defined as
\begin{equation}   \bm{C}_{k,i}=\gamma(k)\cos\!\left(\frac{\pi}{n}\left(i+\frac{1}{2}\right)k\right),
\end{equation}
where $\gamma(k)$ is the normalization factor ensuring orthonormality (\ie, $\bm{CC}^\top=\bm{I}$), given by $\gamma(0)=\sqrt{\frac{1}{n}}$ and $\gamma(k)=\sqrt{\frac{2}{n}}$ for $k\ge 1$, and the half-sample shift $\left(i+\frac{1}{2}\right)$ is the standard DCT-II sampling convention. Then the DCT coefficient matrix $\bm{F}$ of $\bm{I_b}$ is obtained by:
\begin{equation}
    \bm{F}^j = \bm{C}^j \bm{I}_b^j \bm{C}^{j\top},
\end{equation}
where $j$ represents the $j$-th block. During the attack, we perturb only the low-frequency components in each DCT coefficient block, i.e., the top-left $s\times s$ sub-block $\bm{F}_{0:s-1,\,0:s-1}^j$ for each $j$, while keeping the remaining coefficients unchanged. After applying the perturbation, we use the inverse DCT (iDCT) to transform the perturbed coefficient map back to the image space:
\begin{equation}
{\bm{I}_b^j}' = \bm{C}^{j\top}
\begin{bmatrix}
\bm{F}_{0:s-1,\;0:s-1}^j+\bm{\delta}^j & \bm{F}_{0:s-1,\;s:n-1}^j\\[3pt]
\bm{F}_{s:n-1,\;0:s-1}^j & \bm{F}_{s:n-1,\;s:n-1}^j
\end{bmatrix}\bm{C}^j,
\end{equation}
where ${\bm{I}_b^j}'$ is the $j$-th perturbed image block, $\bm{\delta}^j\in\mathbb{R}^{s\times s}$ denotes the low-frequency perturbation. For simplicity, we denote the intermediate block-wise coefficient matrix as $\bm{F}^j_{ss}+\bm{\delta}_{ss}^j$. After concatenating all image blocks, we obtain the final perturbed image $\bm{I}'$. Although we optimize the perturbation in the frequency domain, applying the iDCT projects it back to the pixel space, allowing us to obtain the desired adversarial example. Next, we apply our method to two black-box attack techniques to construct adversarial examples.

\subsubsection{Gradient-based Approach}
We leverage a widely used variant of Natural Evolution Strategies (NES)\cite{ilyas2018black} to estimate gradients for feed-forward 3DGS models, which approximates gradients by averaging loss-weighted perturbations sampled from a noise distribution. The procedure can be formulated as follows:
\begin{equation}
\nabla_{\bm{I}}\mathcal{L}(\theta)
\approx
\frac{1}{2M\sigma}\sum_{m=1}^{M}
(\mathcal{L}(\bm{I}_m'^+,\bm{I}_{r,(m)}'^+)-\mathcal{L}(\bm{I}_m'^-,\bm{I}_{r,(m)}'^-)\mathbf{u}_m,
\end{equation}
where 
\begin{equation}
\label{back to img eq}
    \bm{I}_m'^+={\text{cat}}_{j=1}^J(\bm{C}^{j\top}(\bm{F}_{ss}^j+\sigma\bm{u}_m^j)\bm{C}^j), \bm{I}_m'^-={\text{cat}}_{j=1}^J(\bm{C}^{j\top}(\bm{F}_{ss}^j-\sigma\bm{u}_m^j)\bm{C}^j),
\end{equation}
$\bm{u}_m$ is standard Gaussian noise, $\text{cat}(\cdot)$ denotes the concatenation operation, $\sigma$ is the search variance, $J$ is the number of blocks, and $M$ is the number of noise samples, $\bm{I}_{r,(m)}'^+$ and $\bm{I}_{r,(m)}'^-$ are the corresponding rendered images. Once we obtain the estimated gradient, we can employ it in \cref{pgd eq} to optimize the perturbation. The detailed algorithm is provided in Appendix A.

\subsubsection{Gradient-free Approach}
Although gradient-based methods are intuitive and efficient, they can fail when the model is insensitive to noise or when the gradient is non-smooth. Therefore, we apply our approach to another gradient-free black-box search technique, covariance matrix adaptation evolution strategy (CMA-ES)\cite{hansen2016cma}. CMA-ES iteratively samples candidate perturbations from a multivariate Gaussian distribution and adapts its mean and covariance based on the best-performing samples to guide the search toward better optima. At each iteration $t$, we sample $\mathcal{B}$ candidate perturbations on DCT coefficient:
\begin{equation}
    \bm{\delta}_{\beta}^{j,(t)} \sim \mathcal{N}_\beta\!\bigl(\bm{a}^{(t)},\,\bm{V}^{(t)}\bigr),
\qquad \beta=1,2,\ldots,\mathcal{B},
\end{equation}
where $\bm{a}^{(t)}$ is the mean and $\bm{V}^{(t)}$ is the covariance matrix. Their initial values are set to zero and the identity matrix, respectively. We use the first expression in \cref{back to img eq} to obtain the perturbed image $\bm{I}_\beta'^{(t)}$ ($\bm{I}_\beta'^{(t)}={\text{cat}}_{j=1}^J(\bm{C}^{j\top}(\bm{F}_{ss}^{j,(t)}+\bm{\delta}_{\beta}^{j,(t)})\bm{C}^j)$), and then perform a forward pass through the model to obtain the rendered image $\bm{I}'^{(t)}_{r,(\beta)}$. Then, we compute the loss of the $\mathcal{B}$ candidate pertubed images using \cref{loss eq}:
\begin{equation}
    \mathcal{L}\!\left(\bm{I}_{1:\mathcal{B}}'^{(t)},\bm{I}'^{(t)}_{r,(1:\mathcal{B})}\right)\;\ge\;
\mathcal{L}\!\left(\bm{I}_{2:\mathcal{B}}'^{(t)},\bm{I}'^{(t)}_{r,(2:\mathcal{B})}\right)\;\ge\;\cdots\;\ge\;
\mathcal{L}\!\left(\bm{I}_{\mathcal{B}:\mathcal{B}}'^{(t)},\bm{I}'^{(t)}_{r,(\mathcal{B}:\mathcal{B})}\right),
\end{equation}
sort them in descending order, and select the top $\mathcal{B}/2$ perturbation candidates, which aligns with our attack objective of maximizing the loss, effectively providing a gradient-like search direction without explicit gradients. Next, the mean and covariance are updated by the following formulations:
\begin{gather}
    \bm{a}^{(t+1)} \;=\; \sum_{\beta=1}^{\mathcal{B}/2} w_\beta\,\bm{\delta}^{(t)}_{\beta:\mathcal{B}}, \\
    \bm{V}^{(t+1)}
    = (1-c_{\mu})\,\bm{V}^{(t)}
    + c_{\mu}\sum_{\beta=1}^{\mathcal{B}/2} w_\beta\,
    \bigl(\bm{\delta}^{(t)}_{\beta:\mathcal{B}}-\bm{a}^{(t)}\bigr)
    \bigl(\bm{\delta}^{(t)}_{\beta:\mathcal{B}}-\bm{a}^{(t)}\bigr)^{\top},
\end{gather}
where $c_\mu$ is the learning rate for updating the covariance matrix, $\bm{\delta}^{(t)}_{\beta:\mathcal{B}}={\text{cat}}_{j=1}^J\allowbreak
(\bm{\delta}_{\beta:\mathcal{B}}^{j,(t)})$. $w_\beta$ is the recombination weight of the $\beta$-th ranked perturbation, which is computed by $w'_\beta=ln(\frac{\mathcal{B}+1}{2})-ln(\beta)$ and then normalized by $w_\beta=w'_\beta/\sum_{\beta=1}^{\mathcal{B}/2}w'_\beta$. At final step, the noise sampled from $\mathcal{N}\!\bigl(\bm{a}^{(T)},\,\bm{V}^{(T)}\bigr)$ is the desired perturbation. 

\begin{table}[h]
\centering
\caption{Quantitative comparison of victim model performance with vs. without our attack on the RE10K dataset. GB denotes the gradient-based variant, and GF denotes the gradient-free variant. Percentages measure the degree of performance degradation.}
\label{tab:re10k}
{\fontsize{7pt}{8pt}\selectfont
\begin{tabular}{l ccccc}
\toprule
Victim Models & \cellcolor{gray!20}PSNR $\downarrow$ & \cellcolor{gray!20}SSIM $\downarrow$ & \cellcolor{gray!20}LPIPS $\uparrow$ & \cellcolor{gray!20}CLIP $\downarrow$ & \cellcolor{gray!20}DINO $\downarrow$ \\
\midrule
DepthSplat           & 21.09 & 0.710 & 0.228 & 0.956 & 0.930 \\
DepthSplat+Ours (GB) & 7.57\textcolor[HTML]{B7282E}{\tiny($-64.1\%$)} & 0.289\textcolor[HTML]{B7282E}{\tiny($-59.3\%$)} & 0.581\textcolor[HTML]{B7282E}{\tiny($+154.8\%$)} & 0.740\textcolor[HTML]{B7282E}{\tiny($-22.3\%$)} & 0.492\textcolor[HTML]{B7282E}{\tiny($-47.1\%$)} \\
DepthSplat+Ours (GF) & 9.73\textcolor[HTML]{B7282E}{\tiny($-53.9\%$)}  & 0.437\textcolor[HTML]{B7282E}{\tiny($-38.5\%$)} & 0.514\textcolor[HTML]{B7282E}{\tiny($+125.4\%$)} & 0.803\textcolor[HTML]{B7282E}{\tiny($-16.0\%$)} & 0.727\textcolor[HTML]{B7282E}{\tiny($-21.8\%$)} \\
\midrule
NoPoSplat            & 22.50 & 0.781 & 0.165 & 0.957 & 0.938 \\
NoPoSplat+Ours (GB)  & 17.64\textcolor[HTML]{B7282E}{\tiny($-21.6\%$)} & 0.523\textcolor[HTML]{B7282E}{\tiny($-33.0\%$)} & 0.417\textcolor[HTML]{B7282E}{\tiny($+152.7\%$)} & 0.893\textcolor[HTML]{B7282E}{\tiny($-6.7\%$)} & 0.816\textcolor[HTML]{B7282E}{\tiny($-13.0\%$)} \\
NoPoSplat+Ours (GF)  & 14.02\textcolor[HTML]{B7282E}{\tiny($-37.7\%$)} & 0.397\textcolor[HTML]{B7282E}{\tiny($-49.2\%$)} & 0.549\textcolor[HTML]{B7282E}{\tiny($+232.7\%$)} & 0.856\textcolor[HTML]{B7282E}{\tiny($-10.6\%$)} & 0.729\textcolor[HTML]{B7282E}{\tiny($-22.3\%$)} \\
\midrule
AnySplat             & 18.94 & 0.672 & 0.271 & 0.928 & 0.938 \\
AnySplat+Ours (GB)   & 12.80\textcolor[HTML]{B7282E}{\tiny($-32.4\%$)} & 0.431\textcolor[HTML]{B7282E}{\tiny($-35.9\%$)} & 0.588\textcolor[HTML]{B7282E}{\tiny($+117.0\%$)} & 0.813\textcolor[HTML]{B7282E}{\tiny($-12.4\%$)} & 0.791\textcolor[HTML]{B7282E}{\tiny($-15.7\%$)} \\
AnySplat+Ours (GF)   & 14.49\textcolor[HTML]{B7282E}{\tiny($-23.5\%$)} & 0.460\textcolor[HTML]{B7282E}{\tiny($-31.5\%$)} & 0.561\textcolor[HTML]{B7282E}{\tiny($+107.0\%$)} & 0.859\textcolor[HTML]{B7282E}{\tiny($-7.4\%$)} & 0.832\textcolor[HTML]{B7282E}{\tiny($-11.3\%$)} \\
\bottomrule
\end{tabular}%
}
\end{table}

\begin{table}[h]
\centering
\caption{Quantitative comparison of victim model performance with vs. without our attack on the DL3DV dataset. GB denotes the gradient-based variant, and GF denotes the gradient-free variant. Percentages measure the degree of performance degradation.}
\label{tab:dl3dv}
{\fontsize{7pt}{8pt}\selectfont
\begin{tabular}{l ccccc}
\toprule
Victim Models & \cellcolor{gray!20}PSNR $\downarrow$ & \cellcolor{gray!20}SSIM $\downarrow$ & \cellcolor{gray!20}LPIPS $\uparrow$ & \cellcolor{gray!20}CLIP $\downarrow$ & \cellcolor{gray!20}DINO $\downarrow$ \\
\midrule
DepthSplat           & 22.21 & 0.785 & 0.165 & 0.963 & 0.909 \\
DepthSplat+Ours (GB) & 14.38\textcolor[HTML]{B7282E}{\tiny($-35.3\%$)} & 0.546\textcolor[HTML]{B7282E}{\tiny($-30.4\%$)} & 0.448\textcolor[HTML]{B7282E}{\tiny($+171.5\%$)} & 0.837\textcolor[HTML]{B7282E}{\tiny($-13.1\%$)} & 0.738\textcolor[HTML]{B7282E}{\tiny($-18.8\%$)} \\
DepthSplat+Ours (GF) & 16.59\textcolor[HTML]{B7282E}{\tiny($-25.3\%$)} & 0.588\textcolor[HTML]{B7282E}{\tiny($-25.1\%$)} & 0.396\textcolor[HTML]{B7282E}{\tiny($+140.0\%$)} & 0.851\textcolor[HTML]{B7282E}{\tiny($-11.6\%$)} & 0.798\textcolor[HTML]{B7282E}{\tiny($-12.2\%$)} \\
\midrule
NoPoSplat            & 21.91 & 0.745 & 0.173 & 0.960 & 0.903 \\
NoPoSplat+Ours (GB)  & 19.80\textcolor[HTML]{B7282E}{\tiny($-9.6\%$)} & 0.582\textcolor[HTML]{B7282E}{\tiny($-21.9\%$)} & 0.338\textcolor[HTML]{B7282E}{\tiny($+95.4\%$)} & 0.886\textcolor[HTML]{B7282E}{\tiny($-7.7\%$)} & 0.819\textcolor[HTML]{B7282E}{\tiny($-9.3\%$)} \\
NoPoSplat+Ours (GF)  & 17.17\textcolor[HTML]{B7282E}{\tiny($-21.6\%$)} & 0.425\textcolor[HTML]{B7282E}{\tiny($-43.0\%$)} & 0.439\textcolor[HTML]{B7282E}{\tiny($+153.8\%$)} & 0.858\textcolor[HTML]{B7282E}{\tiny($-10.6\%$)} & 0.722\textcolor[HTML]{B7282E}{\tiny($-20.0\%$)} \\
\midrule
AnySplat             & 19.34 & 0.639 & 0.285 & 0.950 & 0.923 \\
AnySplat+Ours (GB)   & 15.47\textcolor[HTML]{B7282E}{\tiny($-20.0\%$)} & 0.467\textcolor[HTML]{B7282E}{\tiny($-26.9\%$)} & 0.502\textcolor[HTML]{B7282E}{\tiny($+76.1\%$)} & 0.867\textcolor[HTML]{B7282E}{\tiny($-8.7\%$)} & 0.815\textcolor[HTML]{B7282E}{\tiny($-11.7\%$)} \\
AnySplat+Ours (GF)   & 17.52\textcolor[HTML]{B7282E}{\tiny($-9.4\%$)} & 0.497\textcolor[HTML]{B7282E}{\tiny($-22.2\%$)} & 0.481\textcolor[HTML]{B7282E}{\tiny($+68.8\%$)} & 0.876\textcolor[HTML]{B7282E}{\tiny($-7.8\%$)} & 0.839\textcolor[HTML]{B7282E}{\tiny($-9.1\%$)} \\
\bottomrule
\end{tabular}%
}
\end{table}
\section{Experiments}
In this section, we answer the following research questions through extensive experiments: \textbf{RQ1:} How effective are our attack algorithms against different feed-forward 3DGS victim models?
\textbf{RQ2:} Can the DCT-based strategy improve attack efficiency? \textbf{RQ3:} How do different levels of attack strength affect the victim model performance?

\subsection{Experiment Settings}
\subsubsection{Datasets}
We use two benchmark datasets for feed-forward 3DGS, including Re10K\cite{zhou2018stereo} and DL3DV\cite{ling2024dl3dv}. These two datasets cover a diverse set of indoor and outdoor scenes.
\begin{figure}[htbp]
  \centering
  \includegraphics[width=0.8\textwidth]{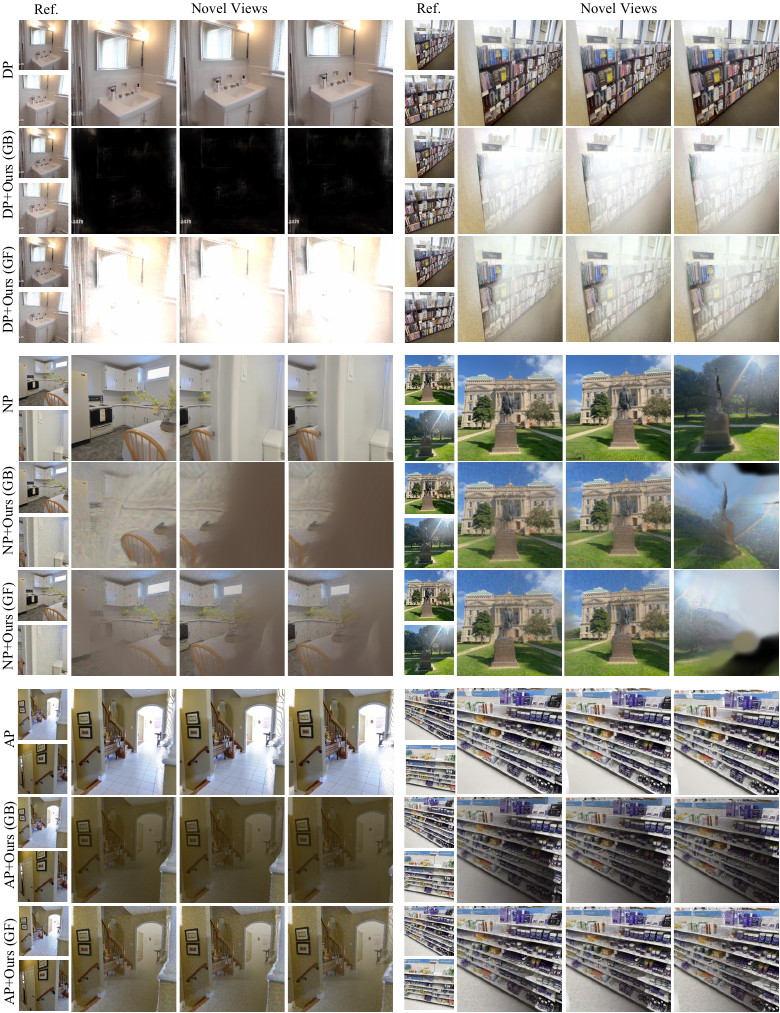}
  \caption{Qualitative results on RE10K and DL3DV ($\epsilon = 8/255$). The left half shows scenes from RE10K, while the right half shows scenes from DL3DV. DP denotes DepthSplat, NP denotes NoPoSplat, and AP denotes AnySplat. GB denotes the gradient-based variant, and GF denotes the gradient-free variant. For each method and each scene, we select two reference views and render three novel views. All baselines use clean inputs, whereas our method uses the optimized adversarial examples as inputs.}
  \label{fig:result1}
\end{figure}
\begin{figure}[htbp]
  \centering
  \includegraphics[width=0.7\textwidth]{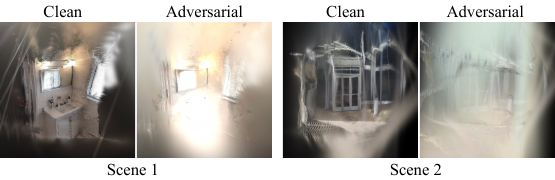}
  \caption{Visualization of 3D Gaussian point clouds.}
  \label{fig:result4}
\end{figure}
\subsubsection{Baselines}
To better validate the effectiveness of our AdvSplat, we focus on feed-forward 3DGS models with sufficiently strong performance. Specifically, we select three representative state-of-the-art models, including one pose-known model, DepthSplat\cite{xu2025depthsplat}, and two  pose-free models NoPoSplat\cite{ye2024no} and AnySplat\cite{jiang2025anysplat}.

\subsubsection{Implementation Details}
We run all models and evaluate our method on a server equipped with four NVIDIA RTX A6000 GPUs. Since we do not know the original training loss used by each black-box model, for fairness and generality, we consistently adopt \cref{loss eq} with $\lambda = 0.05$. We set the image block size $n = 8$ and low frequency block size $s = 3$; accordingly, the number of blocks $J$ is computed as $H\times{W}/n^2$. We set the number of attack iterations to 10,000, the step size $\eta$ to 2/255, while constraining the perturbation magnitude to $\ell_\infty=8/255$. For all baselines, the number of input images $N$ is equal to 2. $M$ and $\mathcal{B}$ are both equal to 40. Notably, to avoid frequent GPU–CPU I/O and enable more efficient computation, \textbf{we implement all AdvSplat operations on CUDA}.

\subsubsection{Metrics}
We adopt standard image quality metrics, including peak signal-to-noise ratio (PSNR), structural similarity (SSIM)\cite{wang2004image}, and learned perceptual image patch similarity (LPIPS)\cite{zhang2018unreasonable}. We additionally include the DINO\cite{zhang2022dino} similarity and the CLIP\cite{radford2021learning} similarity to measure semantic consistency. All metrics are computed between the rendered images and the ground-truth images.

\subsection{Experiment Results}
\subsubsection{Attack Effectiveness of AdvSplat (RQ1)}
We provide a comprehensive evaluation through both quantitative and qualitative results. As shown in \cref{tab:re10k} and \cref{tab:dl3dv}, our attack method significantly degrades the performance of all victim models across all metrics. Notice that, since our objective is the opposite of the reconstruction goal, the interpretation of all metrics is correspondingly reversed: lower is better for PSNR, SSIM, CLIP, and DINO, while higher is better for LPIPS. In general, when gradient estimation is sufficiently accurate, gradient-based methods outperform gradient-free ones, since moving along the gradient direction increases the loss more rapidly. However, in NoPoSplat, the gradient-free variant outperforms the gradient-based variant because NoPoSplat is trained on a mixture of two datasets, resulting in a larger model capacity and making its gradients harder to estimate accurately. \cref{fig:result1} visualizes the input images and the corresponding rendered outputs. As shown, the perturbations on adversarial inputs are nearly imperceptible to human eyes. This is because the $\ell_\infty$ constraint is strictly set to 8/255. Even in the worst case that each pixel is perturbed by at most 8/255, the PSNR between the adversarial and clean images still remains above 30 dB, showing closeness between them. Despite the small perturbation, the quality of the rendered images after attack degrades dramatically, and some results are even completely corrupted. This indicates that, even in the black-box setting, AdvSplat can generate sufficiently strong adversarial examples, exposing the vulnerability of existing feed-forward 3DGS models under such attacks. We also showcase 3D Gaussian point-cloud visualizations for two scenes (due to page limitations) in \cref{fig:result4}. AdvSplat induces drastic changes in the colors and opacities of Gaussians, thereby successfully degrading the rendering quality.

\subsubsection{Efficiency of DCT-based Strategy (RQ2)}
\begin{figure}[tbp]
  \centering
  \includegraphics[width=0.75\textwidth]{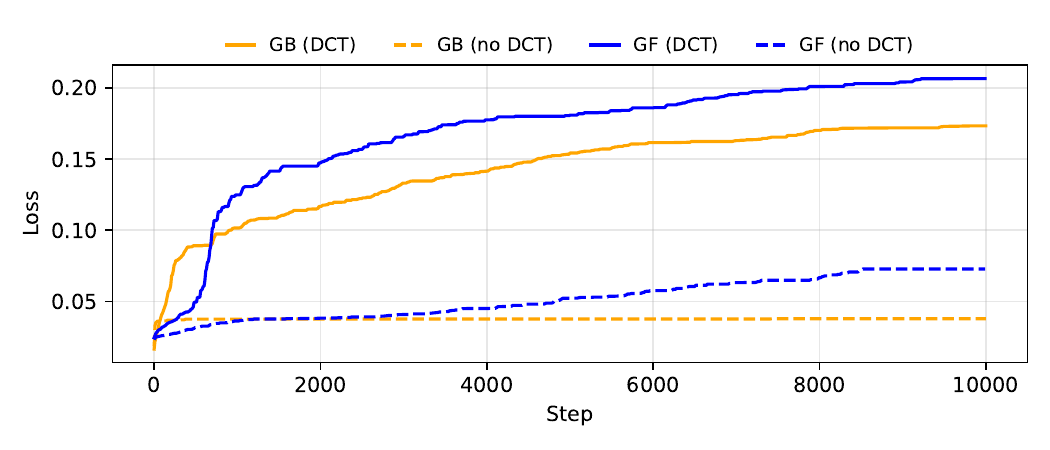}
  \caption{Loss comparison with and without DCT. GB denotes the gradient-based variant, and GF denotes the gradient-free variant.}
  \label{fig:result2}
\end{figure}
Using DCT to sample noise from the low-frequency components in the frequency domain is a key strategy of our method, which lowers the dimensionality of the optimization space. Fewer samples translate into fewer queries to the feed-forward 3DGS model, which not only reduces optimization time but also lowers the cost of querying commercial models. \cref{fig:result2} shows the loss (in  \cref{loss eq}, computed as the average across all victim models and datasets) curves with and without DCT. “no DCT” means sampling noise directly in the full pixel space. For fairness, we use the same number of samples. As shown, for both the gradient-based and gradient-free variants, incorporating DCT and sampling noise in the low-frequency components leads to a higher loss at the same query step once reaching convergence, validating the improved efficiency brought by DCT. 

\begin{table}[h]
\centering
\caption{Victim model performance on the RE10K and DL3DV datasets (We select one scene per dataset and use DepthSplat) under different attack strengths ($\epsilon$ values). GB denotes the gradient-based variant, and GF denotes the gradient-free variant.}
\label{tab:attack_strenths}
{\fontsize{6pt}{7pt}\selectfont
\begin{tabular}{lc ccccc @{\hspace{1em}} ccccc}
\toprule
\multirow{2}{*}{$\epsilon$} & \multirow{2}{*}{Method} & \multicolumn{5}{c}{RE10K} & \multicolumn{5}{c}{DL3DV} \\
\cmidrule(lr){3-7} \cmidrule(l){8-12}
& & \cellcolor{gray!20}PSNR $\downarrow$ & \cellcolor{gray!20}SSIM $\downarrow$ & \cellcolor{gray!20}LPIPS $\uparrow$ & \cellcolor{gray!20}CLIP $\downarrow$ & \cellcolor{gray!20}DINO $\downarrow$ 
& \cellcolor{gray!20}PSNR $\downarrow$ & \cellcolor{gray!20}SSIM $\downarrow$ & \cellcolor{gray!20}LPIPS $\uparrow$ & \cellcolor{gray!20}CLIP $\downarrow$ & \cellcolor{gray!20}DINO $\downarrow$ \\
\midrule
\multirow{2}{*}{2/255} 
& GB & 23.65 & 0.777 & 0.350 & 0.937 & 0.933 & 15.61 & 0.617 & 0.246 & 0.956 & 0.957 \\
& GF & 23.59 & 0.772 & 0.321 & 0.942 & 0.946 & 20.23 & 0.698 & 0.231 & 0.967 & 0.978 \\
\midrule
\multirow{2}{*}{4/255} 
& GB & 5.96  & 0.195 & 0.504 & 0.711 & 0.571 & 9.71  & 0.455 & 0.378 & 0.907 & 0.898 \\
& GF & 9.914 & 0.464 & 0.363 & 0.874 & 0.802 & 14.08 & 0.561 & 0.292 & 0.937 & 0.954 \\
\midrule
\multirow{2}{*}{8/255} 
& GB & 4.38  & 0.050 & 0.663 & 0.554 & 0.060 & 8.09  & 0.384 & 0.493 & 0.858 & 0.726 \\
& GF & 8.39  & 0.608 & 0.548 & 0.701 & 0.908 & 10.82 & 0.456 & 0.406 & 0.854 & 0.847 \\
\midrule
\multirow{2}{*}{16/255} 
& GB & 7.20  & 0.325 & 0.851 & 0.589 & 0.019 & 8.01  & 0.343 & 0.562 & 0.800 & 0.724 \\
& GF & 7.54  & 0.522 & 0.649 & 0.664 & 0.758 & 9.49  & 0.373 & 0.521 & 0.820 & 0.802 \\
\bottomrule
\end{tabular}%
}
\end{table}

\subsubsection{Attack Effectiveness Under Various Attack Strengths (RQ3)}
\begin{figure}[h]
  \centering
  \includegraphics[width=0.8\textwidth]{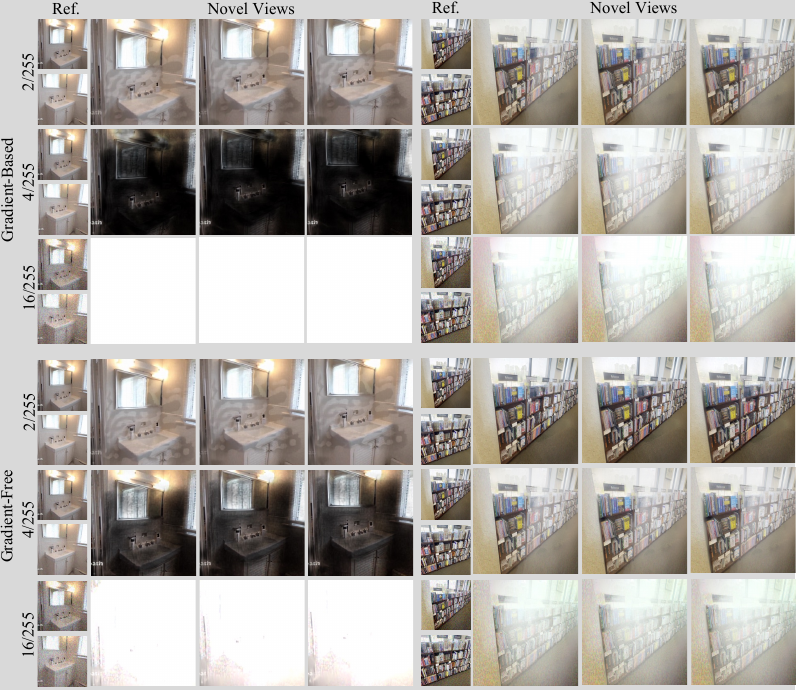}
  \caption{Qualitative results under different attack strengths. The gray background is used to better visualize the all-white rendered images.}
  \label{fig:result3}
\end{figure}
We vary the attack strength, \ie, the $\ell_\infty$ perturbation budget, to further evaluate the effectiveness of our attack. We consider four attack strengths and evaluate both the gradient-based and gradient-free variants on the two datasets. As shown in \cref{tab:attack_strenths}, as the attack strength increases, all metrics exhibit a consistent trend of deterioration. One exception is that on RE10K, for the gradient-based variant, PSNR and SSIM actually improve when the perturbation budget increases from 8/255 to 16/255. However, this does not imply that a stronger attack leads to better rendering quality. As observed in \cref{fig:result1}, when $\epsilon = 8/255$, the rendered images contain many black artifacts. When $\epsilon = 16/255$ (in \cref{fig:result3}), the renderings collapse to nearly all-white images. Since the clean inputs also contain many near-white pixels, this failure case can misleadingly inflate PSNR and SSIM, leading to the observed increase. However, by inspecting the rendered images and considering LPIPS together with the CLIP and DINO similarities, we find that the renderings under $\epsilon = 16/255$ are still of worse quality.

\section{Conclusion}
In this paper, we present the first study on the unexplored robustness issue of feed-forward 3DGS models, and propose AdvSplat, an efficient attack method for them. AdvSplat operates in a black-box setting, requiring only access to inputs and outputs, and performs attacks by sampling perturbations through a frequency-domain parameterization. Our results reveal the vulnerability of feed-forward 3DGS models and highlight new challenges for their commercial deployment. We hope this work will draw the community’s attention and inspire future research on improving the robustness and defenses of these models.
 
\clearpage  


%
%
\bibliographystyle{splncs04}
\bibliography{main}

\clearpage
\appendix
\begin{center}
{\LARGE\textbf{Appendix}}
\end{center}
\section{Algorithm Details}
In the main paper, we present only the core components of our method. Therefore, in the appendix, we provide pseudocode for both the gradient-based and gradient-free variants to help readers better understand the algorithm. Moreover, the gradient-free variant described in the main paper is a simplified version; in this section, we provide the full version in detail.

\begin{algorithm}[h]
\caption{Gradient-Based Frequency-Domain Black-Box Attack}
\label{alg:freq_nes_pgd}
\textbf{Input:} Original image $\bm{I}\in \mathbb{R}^{H \times W \times 3}$, victim feed-forward 3DGS model $f(\cdot)$, adversarial loss $\mathcal{L}$, DCT block size $n$, low-frequency sub-block size $s$, PGD iterations $T$, step size $\eta$, $L_\infty$ perturbation bound $\epsilon$, number of NES noise samples $M$, NES search variance $\sigma$, zero-padding operator $\text{pad}_{s \to n}(\cdot)$ mapping $\mathbb{R}^{s \times s \times 3} \to \mathbb{R}^{n \times n \times 3}$, element-wise sign function $\text{sign}(\cdot)$. \\
\textbf{Output:} Adversarial example $\bm{I}'\in \mathbb{R}^{H \times W \times 3}$
\begin{algorithmic}[1] 
\State Initialize adversarial example: $\bm{I}' \leftarrow \bm{I}$
\For{$t = 0$ \textbf{to} $T-1$}
    \State Divide $\bm{I}'$ into $J$ blocks of equal size $n \times n$: $\{\bm{I}_b^{\prime 0}, \bm{I}_b^{\prime 1}, \dots, \bm{I}_b^{\prime J-1}\}$
    \State Extract current DCT coefficients: $\bm{F}^j = \bm{C}^j \bm{I}_b^{\prime j} \bm{C}^{j\top}, \quad \forall j \in \{0, \dots, J-1\}$
    \State Initialize gradient accumulators: $\mathbf{g}^j = \mathbf{0} \in \mathbb{R}^{s \times s \times 3}, \quad \forall j$
    \vspace{0.05cm}
    \For{$m = 0$ \textbf{to} $M-1$} 
        \State Sample standard Gaussian noise in frequency domain:
        \Statex \hspace{4em}$\mathbf{u}_m^j \sim \mathcal{N}(0, \mathbf{I}_{s \times s \times 3}), \quad \forall j$
        \State Construct perturbed low-frequency coefficients:
        \State \quad $\bm{F}_{ss, m}^{j, +} = \bm{F}_{0:s-1, 0:s-1}^j + \sigma \mathbf{u}_m^j$
        \State \quad $\bm{F}_{ss, m}^{j, -} = \bm{F}_{0:s-1, 0:s-1}^j - \sigma \mathbf{u}_m^j$
        \State Apply iDCT and concatenate to generate probe images $\bm{I}_m^{\prime +}, \bm{I}_m^{\prime -}$ (Eq. (10))
        \State Obtain rendered images: $\bm{I}_{r, (m)}^{\prime +} = f(\bm{I}_m^{\prime +}), \; \bm{I}_{r, (m)}^{\prime -} = f(\bm{I}_m^{\prime -})$
        \State Accumulate gradients: $\mathbf{g}^j \leftarrow \mathbf{g}^j + [\mathcal{L}(\bm{I}_m^{\prime +}, \bm{I}_{r, (m)}^{\prime +}) - \mathcal{L}(\bm{I}_m^{\prime -}, \bm{I}_{r, (m)}^{\prime -})] \mathbf{u}_m^j$
    \EndFor
    \vspace{0.05cm}
    \State Compute spatial block gradients: $\mathbf{g}_{img}^j = \bm{C}^{j\top} \text{pad}_{s \to n}(\mathbf{g}^j) \bm{C}^j, \quad \forall j$
    \State Concatenate all spatial blocks $\mathbf{g}_{img}^j$ to form global spatial gradient 
    \Statex \hspace{3em}$\mathbf{G} \in \mathbb{R}^{H \times W \times 3}$
    \State Estimate global spatial gradient: $\nabla_{\bm{I}'} \mathcal{L} \approx \frac{1}{2 M \sigma} \mathbf{G}$
    \State Sign gradient update in pixel space: $\bm{I}' \leftarrow \bm{I}' + \eta \cdot \text{sign}(\nabla_{\bm{I}'} \mathcal{L})$
    \State Spatial domain $L_\infty$ projection around clean image: $\bm{I}' \leftarrow \text{Clip}(\bm{I}', \bm{I} - \epsilon, \bm{I} + \epsilon)$
    \State Ensure valid pixel range: $\bm{I}' \leftarrow \text{Clip}(\bm{I}', 0, 1)$
\EndFor
\State \Return $\bm{I}'$
\end{algorithmic}
\end{algorithm}

\begin{algorithm}[h]
\caption{Gradient-Free Frequency-Domain Black-Box Attack}
\label{alg:freq_cmaes_full}
\fontsize{9}{9.8}\selectfont
\textbf{Input:} Original image $\bm{I}\in \mathbb{R}^{H \times W \times 3}$, victim model $f(\cdot)$, adversarial loss $\mathcal{L}$, DCT block size $n$, low-frequency subset size $s$, total image blocks $J$, iterations $T$, population size $\mathcal{B}$, initial step size $\sigma_{init}$, $L_\infty$ bound $\epsilon$, zero-padding operator $\text{pad}_{s \to n}(\cdot)$. \\
\textbf{Output:} Adversarial example $\bm{I}'\in \mathbb{R}^{H \times W \times 3}$
\begin{algorithmic}[1] 
\State Let $D = s \times s \times 3$ be the sub-block dimension in the frequency domain, and $D_{total} = J \times D$ be the global dimension
\State Initialize block distributions $\forall j \in \{0, \dots, J-1\}$: mean $\bm{a}^j \leftarrow \mathbf{0} \in \mathbb{R}^D$, diagonal covariance $\bm{V}^j \leftarrow \mathbf{1} \in \mathbb{R}^D$
\State Initialize global step size $\sigma \leftarrow \sigma_{init}$, and block evolution paths $\bm{p}_c^j \leftarrow \mathbf{0} \in \mathbb{R}^D, \bm{p}_s^j \leftarrow \mathbf{0} \in \mathbb{R}^D, \forall j$
\vspace{0.05cm}
\State Set $\mu = \mathcal{B}/2$. Calculate weights: $w'_\beta = \ln(\mu + 0.5) - \ln(\beta + 1)$ and normalize $w_\beta = \frac{w'_\beta}{\sum_{i=0}^{\mu-1} w'_i}$
\State Compute variance effective selection mass: $\mu_{eff} = 1 \big/ \sum_{\beta=0}^{\mu-1} w_\beta^2$
\State Evolution path rates: $c_c = \frac{4 + \mu_{eff}/D_{total}}{D_{total} + 4 + 2\mu_{eff}/D_{total}}$, \quad $c_s = \frac{\mu_{eff} + 2}{D_{total} + \mu_{eff} + 5}$
\State Covariance update rates: $c_1 = \frac{2}{(D_{total} + 1.3)^2 + \mu_{eff}}$, \quad $c_\mu = \min\left(1 - c_1, 2\frac{\mu_{eff} - 2 + 1/\mu_{eff}}{(D_{total} + 2)^2 + \mu_{eff}}\right)$
\State Step-size damping $d_\sigma = 1 + 2\max\left(0, \sqrt{\frac{\mu_{eff} - 1}{D_{total} + 1}} - 1\right) + c_s$, expected norm $\chi_N \approx \sqrt{D_{total}}\left(1 - \frac{1}{4D_{total}} + \frac{1}{21D_{total}^2}\right)$
\vspace{0.05cm}
\For{$t = 0$ \textbf{to} $T-1$}
    \vspace{0.05cm}
    \For{$\beta = 0$ \textbf{to} $\mathcal{B}-1$}
        \State Sample noise and generate candidate sub-blocks: $\bm{z}_\beta^j \sim \mathcal{N}(\mathbf{0}, \bm{I}_D)$ and \Statex \hspace{4em}$\bm{\delta}_\beta^j = \bm{a}^j + \sigma \sqrt{\bm{V}^j} \odot \bm{z}_\beta^j, \;\; \forall j$
        \State Compute spatial blocks via padding \& iDCT: $\mathbf{p}_\beta^j = \bm{C}^{j\top} \text{pad}_{s \to n}(\bm{\delta}_\beta^j) \bm{C}^j, \;\; \forall j$
        \State Concatenate blocks $\mathbf{p}_\beta^j$ to form global spatial perturbation $\mathbf{P}_\beta$
        \State Apply $L_\infty$ and valid pixel range clipping: \State\hspace{1em}$\bm{I}_\beta^{\prime} \leftarrow \text{Clip}(\text{Clip}(\bm{I} + \mathbf{P}_\beta, \bm{I} - \epsilon, \bm{I} + \epsilon), 0, 1)$
        \State Compute fitness: $L_\beta = \mathcal{L}(\bm{I}_\beta^{\prime}, f(\bm{I}_\beta^{\prime}))$
    \EndFor
    \vspace{0.05cm}
    \State Sort descending by fitness: $L_{(0)} \ge L_{(1)} \ge \dots \ge L_{(\mathcal{B}-1)}$ and select top $\mu$ 
    \Statex\hspace{2em}candidates
    \State Update per-block means: $\bm{a}_{old}^j \leftarrow \bm{a}^j$ and $\bm{a}^j \leftarrow \sum_{\beta=0}^{\mu-1} w_\beta \bm{\delta}_{\beta}^j, \;\; \forall j$
    \vspace{0.05cm}
    \State Compute normalized mean differences: $\bm{y}^j \leftarrow (\bm{a}^j - \bm{a}_{old}^j) / \sigma, \;\; \forall j$
    \State Update conjugate paths: $\bm{p}_s^j \leftarrow (1-c_s)\bm{p}_s^j + \sqrt{c_s(2-c_s)\mu_{eff}} (\bm{V}^j)^{-1/2} \odot \bm{y}^j, \;\; \forall j$
    \State Evaluate global Heaviside function $h_\sigma \in \{0, 1\}$ based on the concatenated global 
    \Statex\hspace{2em}path norm $\|\bm{p}_s\|_{global}$ and $\chi_N$
    \State Update covariance paths: $\bm{p}_c^j \leftarrow (1-c_c)\bm{p}_c^j + h_\sigma \sqrt{c_c(2-c_c)\mu_{eff}} \bm{y}^j, \;\; \forall j$
    \State Update global step size: $\sigma \leftarrow \sigma \exp \left( \frac{c_s}{d_\sigma} \left( \frac{\|\bm{p}_s\|_{global}}{\chi_N} - 1 \right) \right)$
    \vspace{0.05cm}
    \State Update diagonal covariance vectors $\bm{V}^j, \;\; \forall j$:
    \State \quad $\bm{V}^j \leftarrow (1 - c_1 - c_\mu)\bm{V}^j + c_1 \big((\bm{p}_c^j)^{\odot 2} + (1-h_\sigma)c_c(2-c_c)\bm{V}^j\big) +$ \Statex\hspace{3em}$c_\mu \sum_{\beta=0}^{\mu-1} w_\beta \big( (\bm{V}^j)^{-1/2} \odot \frac{\bm{\delta}_{\beta}^j - \bm{a}_{old}^j}{\sigma} \big)^{\odot 2}$
\EndFor
\vspace{0.05cm}
\State Extract converged spatial blocks via padding \& iDCT: $\mathbf{p}_{final}^j = \bm{C}^{j\top} \text{pad}_{s \to n}(\bm{a}^j) \bm{C}^j, \;\; \forall j$
\State Concatenate blocks $\mathbf{p}_{final}^j$ to form $\mathbf{P}_{final}$
\State \Return $\bm{I}' \leftarrow \text{Clip}(\text{Clip}(\bm{I} + \mathbf{P}_{final}, \bm{I} - \epsilon, \bm{I} + \epsilon), 0, 1)$
\end{algorithmic}
\end{algorithm}

\clearpage
\section{Additional Results of White-Box Attack}
\begin{figure}[htbp]
  \centering
  \includegraphics[width=0.8\textwidth]{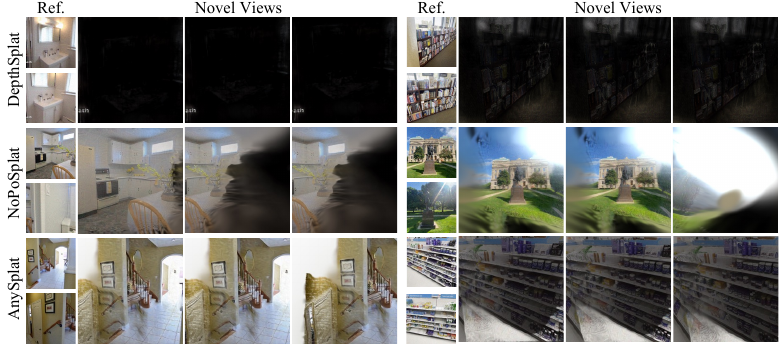}
  \caption{Qualitative results for white-box attack on RE10K and DL3DV ($\epsilon = 8/255$). The left half shows scenes from RE10K, while the right half shows scenes from DL3DV. The clean reconstruction results are shown in \cref{fig:result1}.}
  \label{fig:pgd}
\end{figure}

We provide additional quantitative and qualitative results for the white-box setting in this section. All hyperparameter settings follow the main paper, except that the number of attack iterations is set to 50. Since white-box attacks assume access to model weights and gradients, the resulting renderings are typically worse than the corresponding black-box results reported in the main paper. For pose-free models such as NoPoSplat and AnySplat, evaluating reconstructions requires not only checking for large-scale artifacts but also examining whether the scene coordinate system has shifted. As shown in the second row of the right half of \cref{fig:pgd}, under the target pose, the rendered viewpoint can deviate substantially from the ground-truth viewpoint.

\begin{table}[h]
\centering
\caption{Quantitative comparison of victim model performance with vs. without our white-box attack on the RE10K dataset. Percentages measure the degree of performance degradation.}
\label{tab:re10k_new}
{\fontsize{7pt}{8pt}\selectfont
\begin{tabular}{l ccccc}
\toprule
Victim Models & \cellcolor{gray!20}PSNR $\downarrow$ & \cellcolor{gray!20}SSIM $\downarrow$ & \cellcolor{gray!20}LPIPS $\uparrow$ & \cellcolor{gray!20}CLIP $\downarrow$ & \cellcolor{gray!20}DINO $\downarrow$ \\
\midrule
DepthSplat           & 21.09 & 0.710 & 0.228 & 0.956 & 0.930 \\
DepthSplat+Ours      & 6.82\textcolor[HTML]{B7282E}{\tiny($-67.7\%$)} & 0.223\textcolor[HTML]{B7282E}{\tiny($-68.6\%$)} & 0.594\textcolor[HTML]{B7282E}{\tiny($+160.5\%$)} & 0.677\textcolor[HTML]{B7282E}{\tiny($-29.2\%$)} & 0.370\textcolor[HTML]{B7282E}{\tiny($-60.2\%$)} \\
\midrule
NoPoSplat            & 22.50 & 0.781 & 0.165 & 0.957 & 0.938 \\
NoPoSplat+Ours       & 8.82\textcolor[HTML]{B7282E}{\tiny($-60.8\%$)} & 0.268\textcolor[HTML]{B7282E}{\tiny($-65.7\%$)} & 0.692\textcolor[HTML]{B7282E}{\tiny($+319.4\%$)} & 0.780\textcolor[HTML]{B7282E}{\tiny($-18.5\%$)} & 0.598\textcolor[HTML]{B7282E}{\tiny($-36.2\%$)} \\
\midrule
AnySplat             & 18.94 & 0.672 & 0.271 & 0.928 & 0.938 \\
AnySplat+Ours        & 10.20\textcolor[HTML]{B7282E}{\tiny($-46.1\%$)} & 0.422\textcolor[HTML]{B7282E}{\tiny($-37.2\%$)} & 0.658\textcolor[HTML]{B7282E}{\tiny($+142.8\%$)} & 0.763\textcolor[HTML]{B7282E}{\tiny($-17.8\%$)} & 0.450\textcolor[HTML]{B7282E}{\tiny($-52.0\%$)} \\
\bottomrule
\end{tabular}%
}
\end{table}

\begin{table}[h]
\centering
\caption{Quantitative comparison of victim model performance with vs. without our white-box attack on the DL3DV dataset. Percentages measure the degree of performance degradation.}
\label{tab:dl3dv}
{\fontsize{7pt}{8pt}\selectfont
\begin{tabular}{l ccccc}
\toprule
Victim Models & \cellcolor{gray!20}PSNR $\downarrow$ & \cellcolor{gray!20}SSIM $\downarrow$ & \cellcolor{gray!20}LPIPS $\uparrow$ & \cellcolor{gray!20}CLIP $\downarrow$ & \cellcolor{gray!20}DINO $\downarrow$ \\
\midrule
DepthSplat           & 22.21 & 0.785 & 0.165 & 0.963 & 0.909 \\
DepthSplat+Ours      & 6.54\textcolor[HTML]{B7282E}{\tiny($-70.6\%$)} & 0.202\textcolor[HTML]{B7282E}{\tiny($-74.3\%$)} & 0.625\textcolor[HTML]{B7282E}{\tiny($+278.8\%$)} & 0.665\textcolor[HTML]{B7282E}{\tiny($-30.9\%$)} & 0.205\textcolor[HTML]{B7282E}{\tiny($-77.4\%$)} \\
\midrule
NoPoSplat            & 21.91 & 0.745 & 0.173 & 0.960 & 0.903 \\
NoPoSplat+Ours       & 11.44\textcolor[HTML]{B7282E}{\tiny($-47.8\%$)} & 0.249\textcolor[HTML]{B7282E}{\tiny($-66.6\%$)} & 0.634\textcolor[HTML]{B7282E}{\tiny($+266.5\%$)} & 0.787\textcolor[HTML]{B7282E}{\tiny($-18.0\%$)} & 0.527\textcolor[HTML]{B7282E}{\tiny($-41.6\%$)} \\
\midrule
AnySplat             & 19.34 & 0.639 & 0.285 & 0.950 & 0.923 \\
AnySplat+Ours        & 14.43\textcolor[HTML]{B7282E}{\tiny($-25.4\%$)} & 0.427\textcolor[HTML]{B7282E}{\tiny($-33.2\%$)} & 0.523\textcolor[HTML]{B7282E}{\tiny($+83.5\%$)} & 0.860\textcolor[HTML]{B7282E}{\tiny($-9.5\%$)} & 0.849\textcolor[HTML]{B7282E}{\tiny($-8.0\%$)} \\
\bottomrule
\end{tabular}%
}
\end{table}

\section{Cross Validation of Transfer-Based Attack}
\begin{figure}[H]
  \centering
  \includegraphics[width=0.8\textwidth]{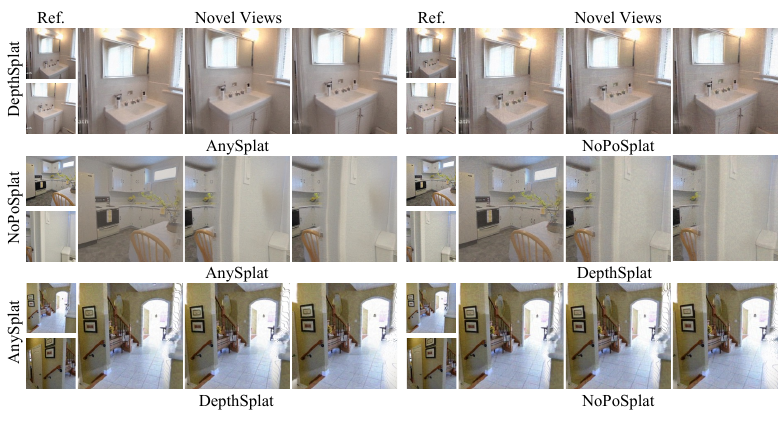}
  \caption{Qualitative results for transfer attack on RE10K ($\epsilon = 8/255$). Each row corresponds to a victim model, and the surrogate model is indicated in each column. The clean reconstruction results are shown in \cref{fig:result1}.}
  \label{fig:transfer}
\end{figure}

In this section, we empirically demonstrate that transfer-based attacks are infeasible for feed-forward 3DGS models. We only use the RE10K dataset because, on DL3DV, the three victim models adopt different preprocessing pipelines; naively unifying image sizes via resizing and cropping would substantially shift the data distribution, making a fair comparison impossible. We perform an exhaustive cross-evaluation over all permutations, \ie using one model as the victim and the other two as surrogate models. Adversarial examples are generated via white-box attacks on the surrogate models, with all other settings kept unchanged. As shown in \cref{tab:transfer_attack}, while transfer-based attacks do degrade reconstruction quality for all victim models, the magnitude is far smaller than that of query-based attacks. \cref{fig:transfer} further shows that, visually, the impact of transfer-based attacks on the rendered images is not pronounced. Overall, transferability does not work well for feed-forward 3DGS models.

\begin{table}[h]
\centering
\caption{Quantitative comparison of transfer attack performance on RE10K. Adversarial examples are generated on a surrogate model and transferred to the victim model. "None" means using the clean data.}
\label{tab:transfer_attack}
\begin{tabular}{cc ccccc}
\toprule
Victim Model & Surrogate Model & \cellcolor{gray!20}PSNR $\downarrow$ & \cellcolor{gray!20}SSIM $\downarrow$ & \cellcolor{gray!20}LPIPS $\uparrow$ & \cellcolor{gray!20}CLIP $\downarrow$ & \cellcolor{gray!20}DINO $\downarrow$ \\
\midrule
\multirow{3}{*}{DepthSplat} & None       & 21.09 & 0.710 & 0.228 & 0.956 & 0.930 \\
                            & NoPoSplat  & 20.32 & 0.653 & 0.406 & 0.927 & 0.889 \\
                            & AnySplat   & 20.10 & 0.658 & 0.294 & 0.947 & 0.905 \\
\midrule
\multirow{3}{*}{NoPoSplat}  & None       & 22.50 & 0.781 & 0.165 & 0.957 & 0.938 \\
                            & DepthSplat & 22.36 & 0.733 & 0.310 & 0.934 & 0.894 \\
                            & AnySplat   & 22.02 & 0.743 & 0.265 & 0.946 & 0.911 \\
\midrule
\multirow{3}{*}{AnySplat}   & None       & 18.94 & 0.672 & 0.271 & 0.928 & 0.938 \\
                            & DepthSplat & 18.15 & 0.580 & 0.440 & 0.894 & 0.894 \\
                            & NoPoSplat  & 17.86 & 0.572 & 0.454 & 0.889 & 0.864 \\
\bottomrule
\end{tabular}%
\end{table}

\clearpage
\end{document}